\documentclass[journal]{IEEEtran}
\IEEEoverridecommandlockouts
\usepackage{cite}
\usepackage{amsmath,amssymb,amsfonts}
\usepackage{algorithmic}
\usepackage{graphicx}
\usepackage{textcomp}
\usepackage{xcolor}
\usepackage{color, colortbl}
\definecolor{Gray}{gray}{0.9}
\definecolor{LightCyan}{rgb}{0.88,1,1}

\usepackage{amsmath}
\usepackage{amssymb}
\usepackage{balance}
\newcommand{\R}{\mathbb{R}}

\def\BibTeX{{\rm B\kern-.05em{\sc i\kern-.025em b}\kern-.08em
    T\kern-.1667em\lower.7ex\hbox{E}\kern-.125emX}}
\begin{document}

\title{\huge Fine-Tuning Llama 2 Large Language Models for Detecting Online Sexual Predatory Chats and Abusive Texts\\
}

\author{Thanh~Thi~Nguyen, 
        Campbell~Wilson,
        and~Janis~Dalins
        % <-this % stops a space
        
\thanks{T. T. Nguyen is with the AiLECS Lab, Faculty of Information Technology, Monash University, Melbourne, Australia e-mail: thanh.nguyen9@monash.edu}% <-this % stops a space

\thanks{C. Wilson is with the AiLECS Lab, Faculty of Information Technology, Monash University, Melbourne, Australia e-mail: campbell.wilson@monash.edu}

\thanks{J. Dalins is with the AiLECS Lab, Australian Federal Police, Melbourne, Australia e-mail: janis.dalins@afp.gov.au}% <-this % stops a space

}

%\author{
%\IEEEauthorblockN{Thanh Thi Nguyen}
%\IEEEauthorblockA{\textit{AiLECS Lab, Faculty of IT} \\
%\textit{Monash University}\\
%Melbourne, Australia \\
%thanh.nguyen9@monash.edu}
%\and
%\IEEEauthorblockN{Campbell Wilson}
%\IEEEauthorblockA{\textit{AiLECS Lab, Faculty of IT} \\
%\textit{Monash University}\\
%Melbourne, Australia \\
%campbell.wilson@monash.edu}
%\and
%\IEEEauthorblockN{Janis Dalins}
%\IEEEauthorblockA{\textit{AiLECS Lab} \\
%\textit{Australian Federal Police}\\
%Melbourne, Australia \\
%janis.dalins@afp.gov.au}
%\and
%\and
%\and

\maketitle

\begin{abstract}
Detecting online sexual predatory behaviours and abusive language on social media platforms has become a critical area of research due to the growing concerns about online safety, especially for vulnerable populations such as children and adolescents. Researchers have been exploring various techniques and approaches to develop effective detection systems that can identify and mitigate these risks. Recent development of large language models (LLMs) has opened a new opportunity to address this problem more effectively. This paper proposes an approach to detection of online sexual predatory chats and abusive language using the open-source pretrained Llama 2 7B-parameter model, recently released by Meta GenAI. We fine-tune the LLM using datasets with different sizes, imbalance degrees, and languages (i.e., English, Roman Urdu and Urdu). Based on the power of LLMs, our approach is generic and automated without a manual search for a synergy between feature extraction and classifier design steps like conventional methods in this domain. Experimental results show a strong performance of the proposed approach, which performs proficiently and consistently across three distinct datasets with five sets of experiments. This study's outcomes indicate that the proposed method can be implemented in real-world applications (even with non-English languages) for flagging sexual predators, offensive or toxic content, hate speech, and discriminatory language in online discussions and comments to maintain respectful internet or digital communities. Furthermore, it can be employed for solving text classification problems with other potential applications such as sentiment analysis, spam and phishing detection, sorting legal documents, fake news detection, language identification, user intent recognition, text-based product categorization, medical record analysis, and resume screening. 
%financial news analysis, 

\end{abstract}

\begin{IEEEkeywords}
large language models, LLMs, Llama 2, fine-tuning, text classification, abusive texts, online sexual predator, cyber grooming, online chats
\end{IEEEkeywords}

\section{Introduction}
\label{sec_int}
Academic research in the realm of detecting online sexual predatory chats and abusive language is diverse and multidisciplinary, spanning machine learning (ML), natural language processing (NLP), psychology, and social sciences. These efforts aim to create safer digital spaces by identifying harmful content and fostering a deeper understanding of the underlying dynamics of online abuse and predatory behaviours. In the fields of ML and NLP, researchers have attempted to develop algorithms capable of identifying predatory behaviours and offensive language in online conversations \cite{milon2022take, khan2023offensive, lykousas2021large, barber2021exposing}. These algorithms often rely on text analysis, pattern recognition, and sentiment analysis by which labelled datasets containing examples of predatory chats or abusive language are used to train ML models \cite{miao2023detecting, chinivar2022online, kavitha2023automatic}. 
%razi2021human

These approaches however do not normally constitute a one-size-fits-all model. This implies that a suggested model could exhibit strong performance on one dataset while potentially demonstrating poor performance on a different dataset. For example, a method may execute competently on English-language text data, but perform imperfectly on non-English text data, or a method may carry out effectively on a balance dataset, but perform deficiently on an imbalance dataset. Likewise, a model may be appropriate for short-text data analysis but may not be suitable for processing and analysing long-text data.

In addition, the existing text classification approaches generally must involve a combination of two main steps, namely feature extraction and classifier design. In order for the entire approach to achieve a satisfactory performance, researchers may need to traverse through an extensive list of combinations involving feature extraction and classification methods to find out the best synchronization or synergy between the two. For example, experiments in~\cite{rezaee2023detecting} were conducted with 16 combinations between four feature engineering methods and four classifiers. The feature engineering methods include variants of the Simple Contrastive Sentence Embedding (SimCSE) such as SimCSE-Base Bert, SimCSE-Large Bert, SimCSE-Base RoBerta and SimCSE-Large RoBerta, whilst a classifier can be either random forest, naïve Bayes, stochastic gradient descent or support vector machine (SVM). 
Similarly, a study in \cite{akhter2021abusive} performed experiments using various ML models such as naïve Bayes, k-nearest neighbors, SVM, logistic regression, rule-based JRip model, convolutional neural network (CNN), long short‑term memory (LSTM), bidirectional LSTM, and convolutional LSTM. Likewise, the work in~\cite{akhter2020automatic} had to experiment with a large number of classifiers, including naïve Bayes, Bayesian network, k-nearest neighbours, Hoeffding tree, J48, reduced error pruning tree, linear SVM, radial SVM, sigmoid SVM, polynomial SVM, random forest, random tree, logistic regression, SimpleLogistic, LogitBoost, OneR and JRip rule-based methods. Regarding the feature extraction/engineering approach, the authors in \cite{akhter2020automatic} encountered various options such as uni-gram, bi-gram, tri-gram, all possible combinations of these n-grams, and they also had to decide whether to use a character n-gram or a word n-gram. 

Recently emerging large language models (LLMs), which constitute a large neural network pretrained on a massive corpus of text data in different languages, have a potential to address the aforementioned issues of the conventional text classification methods. These LLMs have been modelled by billions of parameters and trained on trillions of tokens. For example, the Llama 2 LLMs recently released by Meta GenAI have been trained on two trillion tokens of text data from publicly available sources and have the number of parameters ranging in scale from 7 billion to 70 billion \cite{touvron2023llama}. Another LLM, Generative Pre-trained Transformer 3 (GPT-3), created by OpenAI \cite{brown2020language}, has 175 billion parameters. A newer version of OpenAI's GPT series, i.e., the multimodal GPT-4 model \cite{openai2023gpt4}, could have been probably trained on trillions of tokens and characterized by many more than 175 billion parameters. 
% GPT 4 = 13 trillion tokens and 1.8 trillion parameters
These LLMs could have substantial world knowledge, language understanding, and common-sense reasoning capabilities. With these capabilities, the pretrained LLMs can be extremely beneficial for various tasks. They can, for instance, be further fine-tuned for dialogue use cases (e.g., OpenAI ChatGPT, Meta Llama 2-Chat, Google Bard, and Anthropic Claude 2) or solving text classification problems. 

This paper introduces an approach to text classification problems by fine-tuning the open-source pretrained Llama 2 model with a particular application in detecting online sexual predatory conversations and abusive language. We perform a series of experiments on various datasets and demonstrate a strong performance of the proposed approach against state-of-the-art methods. The next section explains in detail our approach whilst Section \ref{sec_exp_dis} describes datasets and performance metrics used in this study. Section \ref{sec_res_dis} presents experimental results and discussions, followed by concluding remarks and further work in Section \ref{sec_con_fut}.

\section{Llama 2-based Text Classification Approach}
\label{sec_pro_app}
\subsection{Open-source LLMs and Llama 2}

LLMs hold significant potential as proficient AI assistants that can perform sophisticated reasoning tasks in different domains. They have been rapidly and widely adopted among the general public as they can facilitate human interaction through user-friendly chat interfaces. However, the significant computational demands of the LLMs training methodology have constrained their development to only a handful of participants. Popular LLM-based applications are OpenAI ChatGPT, Google Bard, and Anthropic Claude that were built based on closed-source LLMs \cite{touvron2023llama}. Recent open-source LLMs such as BLOOM \cite{scao2022bloom}, LLaMa-1 \cite{touvron2023llama1}, and Falcon \cite{penedo2023refinedweb} are not on par with the closed-source counterparts whose fine-tuning process normally requires significant costs in compute and human annotation to align with human preferences.

\begin{figure}[tb]
\centerline{\includegraphics[width=0.78\linewidth]{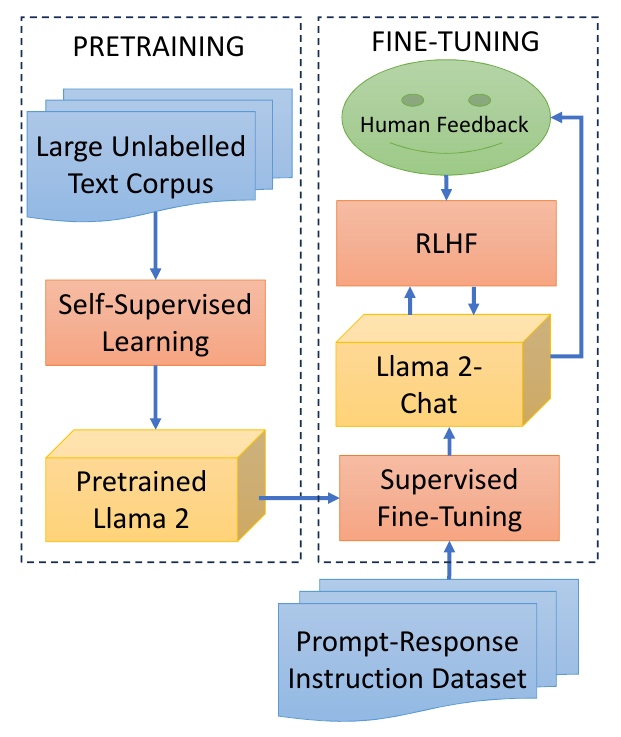}}
\caption{The pretrained Llama 2 model is created by the self-supervised learning using a large unlabelled text corpus while the initial Llama 2-Chat model is built by the supervised fine-tuning using the pretrained models and a small high-quality prompt-response instruction dataset. The initial version of Llama 2-Chat is then iteratively optimized using reinforcement learning with human feedback (RLHF) \cite{touvron2023llama}.}
\label{fig_pretraining}
\end{figure}

Llama 2, an updated version of Llama 1 \cite{touvron2023llama1} and a family of open-source pretrained and fine-tuned LLMs, was recently introduced by Meta GenAI and evaluated as comparable with cutting-edge closed-source LLMs. This creates opportunities for the research community to advance AI in a more transparent way, leading to more responsible development of LLMs. The released Llama 2 family includes pretrained and fine-tuned models, both come with three variants of 7B, 13B, and 70B parameters. The pretrained models were obtained through a self-supervised learning approach using a large text corpus with two trillion tokens. These pretrained models are further optimized based on supervised fine-tuning and reinforcement learning with human feedback (RLHF) using instruction datasets and human-annotated examples for dialogue use cases, resulting in the fine-tuned Llama 2-Chat models. The process to create the pretrained Llama 2 models and fine-tuned Llama 2-Chat models is described in detail in \cite{touvron2023llama}, and summarized in Fig. \ref{fig_pretraining}.

%vocabulary size = 32k tokens
%why we combine these two fields into one

\subsection{Fine-tuning Llama 2 for text classification}

In this study, we use the pretrained Llama 2 7B-parameter model and apply a fine-tuning approach to it for text classification problems. This model along with other Llama 2 variants were built based on an auto-regressive transformer~\cite{vaswani2017attention}. The model architecture and pretraining process involve the use of the RMSNorm pre-normalization \cite{zhang2019root}, the SwiGLU activation function \cite{shazeer2020glu}, and the rotary positional embeddings \cite{su2021roformer}. The LLaMA tokenizer uses the bytepair encoding algorithm \cite{sennrich2016neural} with the implementation from SentencePiece \cite{kudo2018sentencepiece}. The resulting Llama vocabulary has a size of 32k tokens \cite{touvron2023llama}.

\begin{figure}[tbp]
\centerline{\includegraphics[width=\linewidth]{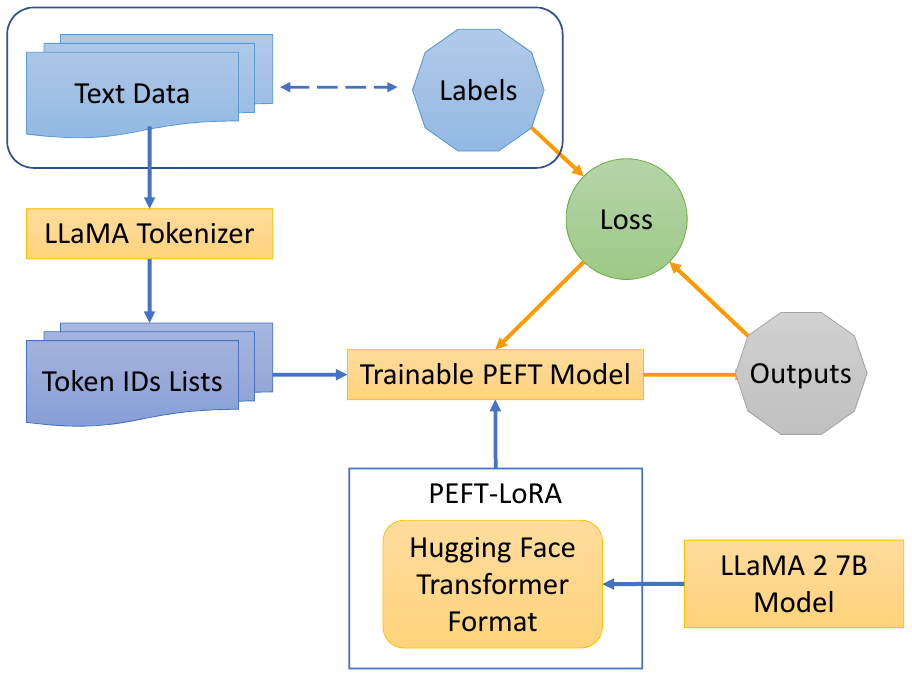}}
\caption{The approach to fine-tuning the pretrained Llama 2 model for text classification problems. Text data are fed into the LLaMA tokenizer and converted to token IDs lists, which are then used as inputs to the trainable PEFT model. This model is created by converting the original LLaMA weights to the Hugging Face transformer model, which is then loaded through the PEFT-LoRA framework.}
\label{fig_method}
\end{figure}

The proposed fine-tuning approach
%for the pretrained Llama 2 model for the text classification purpose 
is depicted in Fig.~\ref{fig_method}. The original weights of the Llama 2 7B-parameter model are converted to the Hugging Face transformer format to take advantage of the Hugging Face fine-tuning tools. The converted model is encapsulated within the PEFT-LoRA framework, which implements the Low-Rank Adaptation (LoRA) method \cite{hu2021lora} based on the Parameter-Efficient Fine-Tuning (PEFT) library \cite{peft}. LoRA is among many PEFT methods, including prefix-tuning \cite{li2021prefix}, prompt-tuning \cite{lester2021power}, Compacter~\cite{karimi2021compacter}, Infused Adapter by Inhibiting and Amplifying Inner Activations - (IA)$^{3}$ \cite{liu2022few}, BitFit \cite{zaken2022bitfit}, SparseAdapter \cite{he2022sparseadapter}, LLaMA-Adapter \cite{zhang2023llama}, AdaLoRA \cite{zhang2023adaptive}, and so on, which are able to deal with the problem of fine-tuning LLMs \cite{lialin2023scaling, ding2023parameter}. The LoRA approach was inspired by the structure-aware intrinsic dimension method \cite{aghajanyan2021intrinsic} and introduces a different way to execute low-rank fine-tuning. In LoRA, the parameter update for a pretrained weight matrix $W_0 \in \R^{d \times k}$ is specified by a product of two low-rank matrices $W_A$ and $W_B$:
\begin{equation}
\delta W = W_A W_B
\end{equation}
where $W_A \in \R^{d \times r}$ and $W_B \in \R^{r \times k}$ are matrices of trainable parameters, and the rank $r \ll \min(d,k)$ \cite{lialin2023scaling}. The pretrained model parameters $W_0$ are frozen during training and do not receive gradient updates. Both $W_0$ and $\delta W = W_A W_B$ are multiplied with the same input \cite{hu2021lora}; hence, the forward pass for $h=W_0 x$ is altered as:
\begin{equation}
h = W_0 x + \delta W x = W_0 x + W_A W_B x
\end{equation}
The matrix $W_A$ is initially set to zero whilst the matrix $W_B$ is started with a random Gaussian initialization. Therefore, $\delta W = W_A W_B$ is zero when the fine-tuning process begins. During training, $\delta W x$ is scaled by $\frac{\alpha}{r}$ where $\alpha$ is a constant in $r$~\cite{hu2021lora}. After training, the trainable parameters can be combined with the original weight matrix $W_0$ by adding the matrix $W_A W_B$ to $W_0$. This approach allows the trainable matrices $W_A$ and $W_B$ to be trained to adapt to the new data while reducing the overall number of updates. As gradients for most of the original pretrained weights do not need to be computed, the GPU memory requirement is reduced significantly, leading to a fast and efficient LoRA-based fine-tuning. 

%Write out the loss function for training ->done
Our fine-tuning approach as presented in Fig. \ref{fig_method} uses a cross-entropy loss function between neural network's output logits $x$ and target $y$, averaged over the mini-batch size $N$:
\begin{equation}
\ell(x, y) = \sum_{n=1}^N \frac{l_n}{\sum_{n=1}^N w_{y_n} }
\end{equation}
%\begin{equation}
%\ell(x, y) = \frac{\sum_{n=1}^N l_n} N
%\end{equation}
where $w$ is a weight vector with each element assigned to each of the classes, and $l_n$ is specified as:
\begin{equation}
l_n = - w_{y_n} \log \frac{\exp(x_{n,y_n})}{\sum_{c=1}^C \exp(x_{n,c})}
\end{equation}
with the logits $x$ consisting of the unnormalized logits for each class, the target $y$ containing class indices, and $C$ being the number of classes. In Fig. \ref{fig_method}, the outputs of the trainable PEFT model correspond to the logits $x$ whilst the labels correspond to the target $y$.
%An equal weight of 1.0 is assigned to each of the classes in this study.

%Include a figure of pretraining using self-supervised learning?
%Draw a figure of methodology pipeline with LoRA fine-tuning,->Done

\section{Datasets and Performance Metrics}
\label{sec_exp_dis}

\subsection{Used Datasets}
We aim to evaluate and validate the proposed fine-tuned Llama 2 text classification approach using datasets with different languages and properties such as data size, imbalance degree, and text length. Experiments in this study are therefore performed using three datasets: the PAN 2012 dataset \cite{inches2012PAN12} for detecting sexual predatory chats, and the Roman Urdu and Urdu datasets \cite{akhter2021abusive,akhter2020automatic} for abusive language detection.

The PAN 2012 competition dataset \cite{inches2012PAN12} was created aiming to serve as a reference point to evaluate performance of various approaches to identify online sexual predators. It is useful for researches in different fields, from text mining, ML to information retrieval and NLP. The dataset contains hundred of thousands of conversations with three types of data, described in detail in \cite{inches2012overview} and summarized as below.

The number of \textit{true positives} is small and they were collected from the PJ website (http://www.perverted-justice.com/). That website holds online chats between convicted sexual predators and volunteers pretending to be underage teenagers. The number of \textit{false positives} (individuals chatting about sex or engaging in a shared topic with the “sexual predator”) is large and they were extracted from Omegle (www.omegle.com). The third data type comprises a large number of \textit{false negatives} (general chats between individuals on different topics), which were collected from the website of the IRC channel managers: http://www.irclog.org/ and http://krijnhoetmer.nl/irclogs/. 

\begin{table*}[htbp]
\caption{Comparisons between Our PAN12 Preprocessed Datasets and Those in \cite{borj2020preprocessing, borj2021detecting, rezaee2023detecting}}
\begin{center}
\begin{tabular}{|c|c|c|c|c|c|c|}
\hline
\textbf{Preprocessed}&\multicolumn{3}{c|}{\textbf{Those in \cite{borj2020preprocessing, borj2021detecting, rezaee2023detecting}}} &\multicolumn{3}{c|}{\textbf{Ours}}\\
\cline{2-7} 
\textbf{PAN12 Datasets} & \textbf{\textit{Predatory}} & \textbf{\textit{Non-Predatory}} & \textbf{\textit{Imbalance (\%)}} & \textbf{\textit{Predatory}} & \textbf{\textit{Non-Predatory}} & \textbf{\textit{Imbalance (\%)}}\\
\hline
Training & 951 & 8,477 & \textbf{11.22} & 952 & 8,522 & \textbf{11.17} \\
\hline
Testing & 1,697 & 19,922 & \textbf{8.52} & 1,698 & 20,024 & \textbf{8.48} \\
\hline
\end{tabular}
\label{tab_pan12_prepro}
\end{center}
\end{table*}

The most recent studies using this dataset \cite{borj2021detecting, borj2020preprocessing, rezaee2023detecting} had preprocessed the conversations to eliminate irrelevant entries. Specifically, in the first step, they eliminated all conversations having only one author or more than two authors because a predatory conversation normally has two authors. In the second step, conversations with less than 7 messages were discarded because they did not contain sufficient information to be classified as either predatory or non-predatory. In the third step, they eliminated non-English words that did not have any special meaning or did not use standard grammar, had many slang and emoticons because those kinds of non-meaningful words and symbols did not contain any useful information for training. %This helped them to evaluate the real impact of the feature engineering step and classification step in their approach pipeline.
To avoid subjectivity and ensure our results are reproducible, we follow the first and second steps but did not follow the third step. Our training and testing data after preprocessing are therefore noisier and more challenging than the preprocessed data in \cite{borj2021detecting, borj2020preprocessing, rezaee2023detecting}. Table \ref{tab_pan12_prepro} details the statistical differences between our PAN12 preprocessed results and those in the mentioned studies. 
%The third preprocessing step in \cite{borj2021detecting, borj2020preprocessing, rezaee2023detecting} is too subjective that we could not reproduce exactly the same preprocessed training and testing datasets as theirs. 
It is worth noting that our preprocessed datasets are more imbalanced as the imbalance degree of the preprocessed datasets in \cite{borj2020preprocessing, borj2021detecting, rezaee2023detecting} is 11.22\% for training and 8.52\% for testing whilst those numbers of ours are 11.17\% and 8.48\%, respectively.

The Roman Urdu dataset contains 147,180 comments on multiple YouTube videos and publicly available at: https://github.com/shaheerakr/roman-urdu-abusive-comment-detector. The comments were in the Roman Urdu language and labelled as abusive or non-abusive. Similar to the previous works in \cite{akhter2021abusive,akhter2020automatic}, ten thousand data points consisting of 5,000 abusive (i.e., positive class) and 5,000 non-abusive (i.e., negative class) comments were randomly extracted from this dataset to validate our proposed method. 

The Urdu dataset was created by manually collecting YouTube comments in Urdu. These comments were made to videos in different topics from political to entertainment, sports, and religion. This dataset is available at: https://github.com/pervezbcs/Urdu-Abusive-Dataset with 2,170 comments, which were annotated by local speakers as either abusive or non-abusive. 

\begin{figure}[bp]
\centerline{\includegraphics[width=\linewidth]{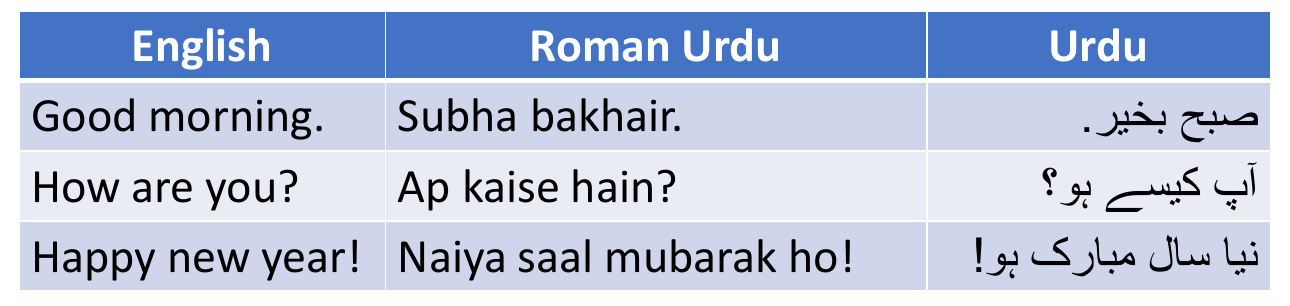}}
\caption{A comparison between English, Roman Urdu, and Urdu via several example sentences.}
\label{fig_languages}
\end{figure}

Urdu and Roman Urdu are two distinct writing systems used to represent the same language (i.e., Urdu), which is primarily spoken in Pakistan and parts of India. Urdu script is a unique Arabic script with extra letters and characters, written from right to left. On the other hand, Roman Urdu is a phonetic representation of Urdu using the Latin alphabet. It allows Urdu speakers to communicate using English characters, making it more accessible, especially in digital communications. Several examples of English, Roman Urdu, and Urdu sentences are presented in Fig. \ref{fig_languages}, which show distinct characteristics of these languages. It would be challenging to build a text classification model that could perform effectively across the languages. 
%Add a figure of examples Good Morning. How are you? Happy New Year. Alphabets of Urdu language in the Nastaleeq script -> Done

\begin{table}[htbp]
\caption{Statistics of the Experimental Datasets}
\begin{center}
\begin{tabular}{|c|c|c|c|}
\hline
\textbf{Dataset}&\multicolumn{3}{c|}{\textbf{Dataset Names}} \\
\cline{2-4} 
\textbf{Properties} & \textbf{\textit{PAN12}}& \textbf{\textit{Roman Urdu}}& \textbf{\textit{Urdu}} \\
\hline
Total no. of data samples/points & 31,196 & 10,000 & 2,170 \\
\hline
No. of positive-class data points & 2,650 & 5,000 & 1,108 \\
\hline
No. of negative-class data points & 28,546 & 5,000 & 1,062 \\
\hline
Minimum length of a sequence & 1 & 1 & 2 \\
\hline
Maximum length of a sequence & 122,763 & 511 & 37 \\
\hline
No. of data points for training & 9,474 & 8,000$^{*}$ & 1,736$^{*}$ \\
\hline
No. of data points for testing & 21,722 & 2,000$^{*}$ & 434$^{*}$ \\
\hline
\multicolumn{4}{l}{$^{*}$Corresponding to the ratio 80:20, but the 90:10 ratio is also used.}
\end{tabular}
\label{tab_data_stats}
\end{center}
\end{table}

A statistical summary of all three experimental datasets is presented in Table \ref{tab_data_stats}. The PAN12 dataset is highly imbalanced with the number of positive-class data points (i.e., predatory conversations) contributing just around 8.49\% to the total number of data samples. The Roman Urdu and Urdu datasets are balanced with the number of data samples in two classes being approximately equal.

\subsection{Performance Metrics}
The following metrics are used to measure performance of competing methods: accuracy, true positive rate (TPR), false positive rate (FPR), and $F_\beta$ with $\beta$ being equal to 0.5 (i.e., $F_{0.5}$ score) and 1.0 (i.e., $F_1$ score or  $F$-measure).
\begin{equation}
\text{Accuracy} = \frac{TP + TN}{TP + TN + FP + FN}
\end{equation}
where $TP$ refers to the number of true positive predictions, $TN$ refers to the number of true negative predictions, $FP$ refers to the number of false positive predictions, and $FN$ refers to the number of false negative predictions.

\begin{equation}
\text{TPR} = \frac{TP}{TP + FN}
\end{equation}

\begin{equation}
\text{FPR} = \frac{FP}{FP + TN}
\end{equation}

%The F0.5-score (to emphasis the importance of fewer false positives) and the F2-score (to emphasis the importance of fewer false negatives).

\begin{equation}
F_{\beta} = (1 + \beta^2) \cdot \frac{\text{Precision} \cdot \text{Recall}}{(\beta^2 \cdot \text{Precision}) + \text{Recall}}
\end{equation}
where $\beta$ is a parameter that determines the relative importance of Precision and Recall. The $F_1$ score gives equal weights to Precision and Recall whilst the $F_{0.5}$ score emphasizes Precision, putting more attention on minimizing false positives rather than minimizing false negatives. %useful to highlight the importance of reducing false positives. 

\section{Results and Discussions}
\label{sec_res_dis}
With the aim towards a one-size-fits-all method, we set the same value across all experiments for each of the fine-tuning parameters in this study. The LoRA attention dimension (the rank of the update matrices) is set to 8, whilst the LoRA alpha (the alpha parameter for LoRA scaling) is equal to 16. The resulting number of trainable parameters is 4,210,688, which is approximately 0.064\% of the total of 6,611,554,304 parameters.
%r=8: the rank of the update matrices, expressed in int. Lower rank results in smaller update matrices with fewer trainable parameters. %r: The dimension used by the LoRA update matrices.
The LoRA dropout (the dropout probability for LoRA layers) is set to 0.1. The fine-tuning process uses the AdamW optimizer \cite{loshchilov2018decoupled} with a learning rate of $2 \times 10^{-5}$ and an epsilon value of $10^{-8}$, and runs through 20 epochs for all experiments.

\begin{figure}[tbp]
\centerline{\includegraphics[width=\linewidth]{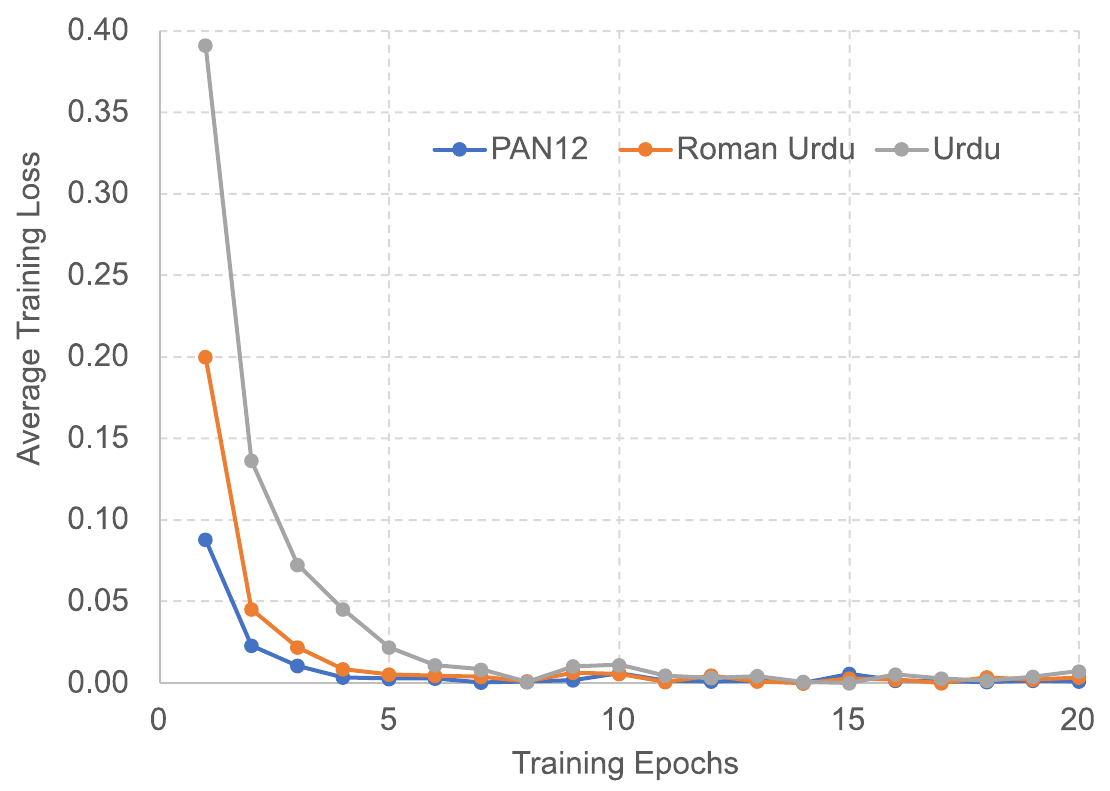}}
\caption{Average training loss for each epoch while we fine-tune the Llama2 7B model using each of the three experimental datasets (i.e., PAN12, Roman Urdu, and Urdu) with 20 epochs.}
\label{fig_train_loss}
\end{figure}

\begin{figure}[tbp]
\centerline{\includegraphics[width=\linewidth]{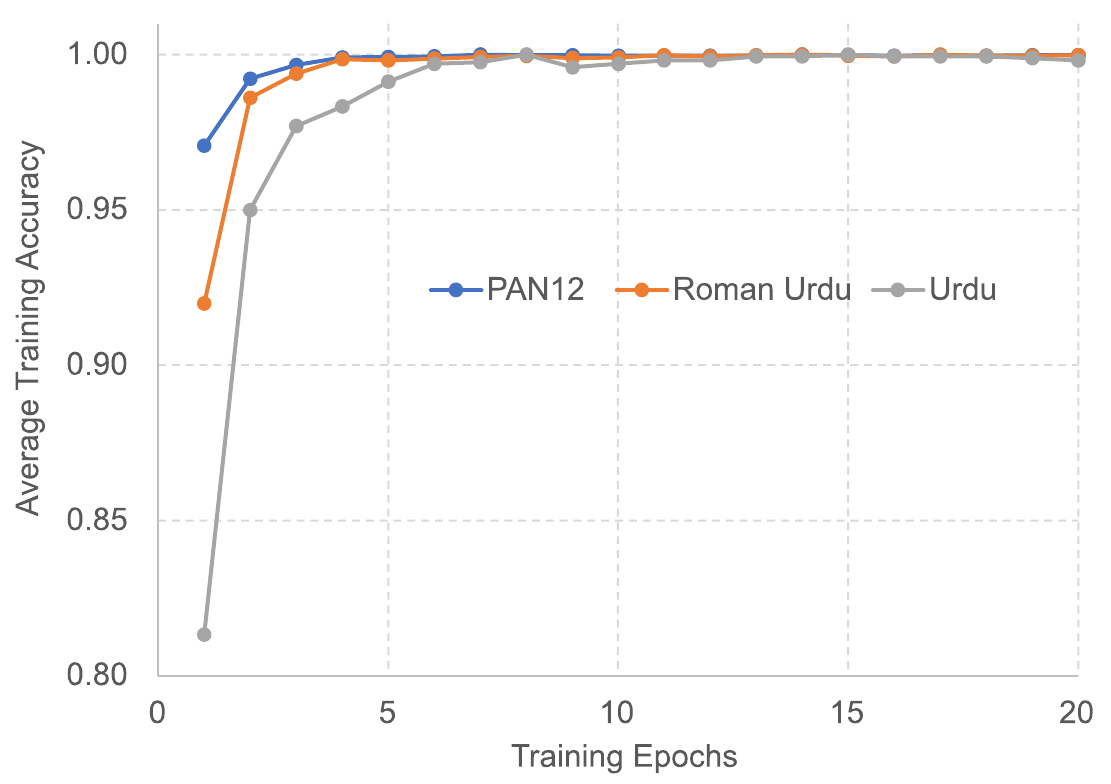}}
\caption{Average training accuracy for each epoch. Among three experimental datasets, PAN12 has the largest number of training data points (i.e., 9,474), whilst Roman Urdu has 8,000, and Urdu has the smallest number (i.e., 1,736).}
\label{fig_train_acc}
\end{figure}

Figs. \ref{fig_train_loss} and \ref{fig_train_acc} show the average training loss and accuracy for each epoch. The largest and also English-language dataset, i.e., PAN12, produces the smallest training loss and largest accuracy after the first epoch, whilst the smallest and also non-English dataset, i.e., Urdu, has the largest training loss and smallest accuracy after the first epoch. This is understandable because Llama 2 was pretrained mostly on the English-language corpus, with a very small fraction of other languages. It is worth noting that both Urdu and Roman Urdu languages could have been used for Llama 2 pretraining with less than 0.005\% \cite{touvron2023llama}, which is a considerably small percentage. Meta GenAI warned that the Llama 2 models may not be suitable for use in other languages because the majority of the pretraining corpus is in English \cite{touvron2023llama}. With the fine-tuning approach to text classification problems proposed in this study, we found that the Llama 2-based approach performed proficiently not only on the English data but also on the Roman Urdu and Urdu data, as showcased in the next subsections.

\subsection{Sexual predatory chat detection using PAN12 dataset}
A comparative summary of the performance metrics for different methods used in detecting online sexual predatory chats is displayed in Table \ref{tab_predator_res}. The methods are evaluated based on their accuracy, $F_1$ score, and $F_{0.5}$ score, which provide insights into their effectiveness. The most recent work on this topic, i.e., Borj et al. \cite{rezaee2023detecting}, achieved a high accuracy of 0.99, an $F_1$ score of 0.97, and an $F_{0.5}$ score of 0.98 (i.e., values in the row ``Borj et al. \cite{rezaee2023detecting} – Fusion" in Table \ref{tab_predator_res}). Our method based on Llama 2 fine-tuning stands out with excellent accuracy with a rounded value of 1.00, and strong $F_1$ and $F_{0.5}$ scores both of 0.98, although it has to deal with noisier and more imbalanced training and testing datasets compared with the studies in Borj et al. \cite{borj2021detecting, rezaee2023detecting}. %This indicates its superior performance in detecting online sexual predatory conversations. 

The performance measures reported in Table \ref{tab_predator_res} for the method in \cite{rezaee2023detecting} (row ``Borj et al. \cite{rezaee2023detecting} – Fusion") are the best result of trialling three fusion approaches at both score-level fusion and decision-level fusion (i.e., sum rule, product rule, and majority voting) on outcomes of 16 individual models. The best performance for an individual model is actually that of the combination of SimCSE-Base RoBerta and SVM, which resulted in an accuracy of 0.99, whilst $F_1$ and $F_{0.5}$ scores are both of 0.96 (i.e., values in the row ``Borj et al. \cite{rezaee2023detecting} – Individual" in Table \ref{tab_predator_res}). In contrast to the method in \cite{rezaee2023detecting}, our approach does not have to exhaust various models or all possible combinations of those models. Based on the power of Llama 2 LLMs, our method is deemed a one-shot approach without an exhaustive search among various options.
%that is deemed a one-size-fits-all approach. 

\begin{table}[bp]
\caption{Comparisons between Our Method and Others for Detecting Online Sexual Predatory Chats using the PAN12 Dataset}
\begin{center}
\begin{tabular}{|c|c|c|c|}
\hline
\textbf{Competing}&\multicolumn{3}{c|}{\textbf{Performance Metrics}} \\
\cline{2-4} 
\textbf{Methods} & \textbf{\textit{Accuracy}}& \textbf{\textit{$F_1$}} & \textbf{\textit{$F_{0.5}$}} \\
\hline
Villatoro-Tello et al. \cite{villatoro2012two} & 0.92 & 0.87 & 0.93 \\
\hline
Fauzi \& Bours \cite{fauzi2020ensemble} & 0.95 & 0.90 & 0.93 \\
\hline
Bogdanova et al. \cite{bogdanova2012impact} &  0.97 & - & -  \\
\hline
Borj \& Bours \cite{borj2019predatory} & 0.98 & 0.86 & - \\
\hline
Bours and Kulsrud \cite{bours2019detection} & - & 0.94 & 0.97 \\
\hline
Borj et al. \cite{borj2020preprocessing} & 0.99 & 0.96 & - \\
\hline
Ebrahimi et al. \cite{ebrahimi2016detecting} & - & 0.80 & - \\
\hline
Ebrahimi et al. \cite{ebrahimi2016recognizing} & 0.99 & 0.77 & - \\
\hline
Borj et al. \cite{borj2021detecting} & 0.99 & \textbf{0.99} & 0.94 \\
\hline
Borj et al. \cite{rezaee2023detecting} - Individual & 0.99 & 0.96 & 0.96 \\
\hline
Borj et al. \cite{rezaee2023detecting} - Fusion & 0.99 & 0.97 & 0.98 \\
\hline
\rowcolor{LightCyan}
Our Llama 2 Approach & \textbf{1.00} & \textbf{0.98} & \textbf{0.98} \\
\hline
%\multicolumn{4}{l}{$^{\mathrm{a}}$Example text here}
\end{tabular}
\label{tab_predator_res}
\end{center}
\end{table}

\begin{table*}[ht]
\caption{Comparisons between Our Method and Others for Detecting Abusive Texts using the Roman Urdu and Urdu Datasets}
\begin{center}
\begin{tabular}{|c|c|c|c|c|c|c|c|c|}
%\hline
%\textbf{Datasets}&\multicolumn{4}{c|}{\textbf{Roman Urdu}}&\multicolumn{4}{c|}{\textbf{Urdu}} \\
%\cline{1-9} 
%\multicolumn{9}{c}{\textbf{\textit{}}} \\
\multicolumn{9}{c}{\textbf{Training-Testing Data Ratio (\%): 80-20}} \\
\hline
\textbf{Datasets}&\multicolumn{4}{c|}{\textbf{Roman Urdu}}&\multicolumn{4}{c|}{\textbf{Urdu}} \\
\hline
\textbf{Methods} & \textbf{\textit{Accuracy}} & \textbf{\textit{$F_1$}} & \textbf{\textit{TPR}}& \textbf{\textit{FPR}} & \textbf{\textit{Accuracy}} & \textbf{\textit{$F_1$}} & \textbf{\textit{TPR}}& \textbf{\textit{FPR}} \\
\hline
Akhter et al. \cite{akhter2021abusive} 2-Layer LSTM & - & 85.7 & 85.7 & 14.3 & - & 91.1 & 91.1 & 8.9 \\
\hline
Akhter et al. \cite{akhter2021abusive} 2-Layer BLSTM & - & 86.2 & 85.8 & 14.2 & - & 92.1 & 92.1 & 7.9 \\
\hline
Akhter et al. \cite{akhter2021abusive} 1-Layer LSTM & - & 88.2 & 87.6 & 12.4 & - & 93.5 & 93.5 & 6.5\\
\hline
Akhter et al. \cite{akhter2021abusive} CLSTM & - & 88.6 & 88.1 & 11.9 & - & 94.3 & 94.2 & 5.8 \\
\hline
\rowcolor{Gray}
Akhter et al. \cite{akhter2021abusive} CNN & - & \textbf{91.6} & 91.4 & 8.6 & - & \textbf{96.2} & 96.1 & \textbf{3.9} \\
\hline
\rowcolor{LightCyan}
Our Llama 2 Approach & \textbf{99.3} & \textbf{99.3} & \textbf{98.9} & \textbf{0.4} & \textbf{95.9} & \textbf{95.9} & \textbf{96.4} & 4.7\\
\hline
\multicolumn{9}{c}{\textbf{\textit{}}} \\

\multicolumn{9}{c}{\textbf{Training-Testing Data Ratio (\%): 90-10}} \\
\hline
\textbf{Datasets}&\multicolumn{4}{c|}{\textbf{Roman Urdu}}&\multicolumn{4}{c|}{\textbf{Urdu}} \\
\hline
\textbf{Methods} & \textbf{\textit{Accuracy}} & \textbf{\textit{$F_1$}} & \textbf{\textit{TPR}}& \textbf{\textit{FPR}} & \textbf{\textit{Accuracy}} & \textbf{\textit{$F_1$}} & \textbf{\textit{TPR}}& \textbf{\textit{FPR}} \\
\hline
Akhter et al. \cite{akhter2020automatic} SVM-Polynomial & 97.7 & 97.7 & - & - & 95.5 & 95.5 & - & - \\
\hline
Akhter et al. \cite{akhter2020automatic} Rule-JRip & 98.2 & 98.2 & - & - & 92.8 & 92.8 & - & -\\
\hline
\rowcolor{Gray}
Akhter et al. \cite{akhter2020automatic} SimpleLogistic & 98.3 & 98.3 & - & - & \textbf{95.9} & \textbf{95.9} & - & - \\
\hline
Akhter et al. \cite{akhter2020automatic} REPTree & 98.9 & 98.9 & - & - & 94.9 & 94.9 & - & - \\
\hline
\rowcolor{Gray}
Akhter et al. \cite{akhter2020automatic} LogitBoost & \textbf{99.2} & \textbf{99.2} & - & - & 94.9 & 94.9 & - & - \\
\hline
\rowcolor{LightCyan}
Our Llama 2 Approach & \textbf{99.5} & \textbf{99.5} & \textbf{99.2} & \textbf{0.2} & \textbf{96.3} & \textbf{96.6} & \textbf{97.4} & 5.0\\
\hline
%\multicolumn{4}{l}{$^{\mathrm{a}}$Example text here}
\end{tabular}
\label{tab_roman_urdu}
\end{center}
\end{table*}

Another method performing highly in this domain is the work in \cite{borj2021detecting}, which achieved an accuracy of 0.99, $F_1$ and $F_{0.5}$ scores of 0.99 and 0.94, respectively. That study attempted to address the imbalance nature of the PAN12 dataset. Because predatory data are scarcer than the non-predatory data, proposed algorithms normally must deal with the data imbalance. The work in \cite{borj2021detecting} introduced a hybrid sampling and class re-distribution method to augment the data. Along with the augmentation process, the data are also perturbed iteratively to enhance the diversity of classifiers and features in an ensemble approach. Although that method obtained an $F_1$ score of 0.99, its $F_{0.5}$ score was just 0.94. The lower $F_{0.5}$ score means that more false positives are flagged. This is not ideal in predator detection and less applicable in law enforcement because it may overwhelm the police agents with so many possible suspects rather than help to optimize their time towards the ``right" suspect \cite{inches2012overview}. Compared with results in~\cite{borj2021detecting}, our method has lower $F_1$ score (0.98 vs 0.99) but higher accuracy (1.00 vs 0.99), and more importantly, higher $F_{0.5}$ score (0.98 vs 0.94), which is more applicable in the law enforcement practice.

\subsection{Abusive language detection using Roman Urdu and Urdu datasets}

In this subsection, we report results of four sets of experiments. For each of the two abusive text datasets (i.e., Roman Urdu and Urdu), we carried out two sets of experiments: one for the training-testing data split ratio of 80-20, and one for the ratio of 90-10. This is because there exist two benchmark studies using these datasets: one in \cite{akhter2021abusive} using the split ratio 80-20 and another in \cite{akhter2020automatic} using the ten-fold cross validation method, i.e., the 90-10 split ratio. %As applying a cross-validation method to the LLM fine-tuning process is too computationally expensive, we instead apply the training-testing split ratio of 90-10 to compare our results with those in \cite{akhter2020automatic}.
We set the same seed number across all four experiments for the random state parameter when splitting the data into training and testing sets.

Five sets of results quoted for \cite{akhter2021abusive} and \cite{akhter2020automatic} in Table \ref{tab_roman_urdu} are of the best five models in each of these studies. The study in \cite{akhter2021abusive} actually attempted 11 different methods, whilst the study in \cite{akhter2020automatic} tried out 17 approaches to search for their best approaches. 

The work in \cite{akhter2021abusive} used the term frequency-inverse document frequency (TF-IDF) and Bag-of-Word as feature engineering approaches to extract features before feeding them into conventional ML models (e.g., NB, SVM, IBK, Logistic, and JRip) or deep learning models (e.g., CNN, 1-Layer LSTM, 2-Layer LSTM, 1-Layer BLSTM, 2-Layer BLSTM and CLSTM). 
Their best approach is the CNN model, which has an $F_1$ score of 91.6\% on the Roman Urdu dataset and 96.2\% on the Urdu dataset. Our Llama 2 approach has an $F_1$ score of 95.9\% on the Urdu dataset, which is 0.3\% lower than that of the best model (i.e., CNN) in \cite{akhter2021abusive}. However, our method is significantly (with 7.7\%) better than their best model on the Roman Urdu dataset. With this dataset, our method obtained a considerably low FPR of 0.4, which is 8.2\% lower than the best result in \cite{akhter2021abusive}. Our TPR results are better than the best in \cite{akhter2021abusive} across both datasets: 98.9\% vs 91.4\% in the Roman Urdu dataset, and 96.4\% vs 96.1\% in the Urdu dataset.

Our method is marginally inferior to the best method in~\cite{akhter2021abusive} on the Urdu dataset because the training set has only 1,736 data points, which is too small to fine-tune the Llama 2 7B-parameter LLM. When the training set is slightly larger, (i.e., in the 90:10 training-testing data ratio with 1,953 training data points, presented in the second part of Table \ref{tab_roman_urdu}), our method performed best on the Urdu dataset. Specifically, it obtained an accuracy of 96.3\% and an $F_1$ score of 96.6\% on the Urdu dataset, surpassing the best results in both \cite{akhter2021abusive} and~\cite{akhter2020automatic}. In the much larger dataset, i.e., Roman Urdu, either with 8,000 or 9,000 training data points, our method consistently outperformed the best results in both \cite{akhter2021abusive} and \cite{akhter2020automatic}. 

Across all four experiments with results presented in Table \ref{tab_roman_urdu}, our method demonstrates a stable and consistent performance compared with state-of-the-art methods in \cite{akhter2021abusive, akhter2020automatic}. For example, our method performs best across the two datasets in the experiments using the 90:10 training-testing data split ratio, i.e., the bottom part of Table \ref{tab_roman_urdu}. On the other hand, the best method in \cite{akhter2020automatic} on the Roman Urdu dataset is the LogitBoost model. This method however is not the best method on the Urdu dataset in \cite{akhter2020automatic} (which is the SimpleLogistic model). Furthermore, the CNN model in \cite{akhter2021abusive} demonstrates good performance on the Urdu dataset but performs poorly on the Roman Urdu dataset compared with most of the models in \cite{akhter2020automatic}. Specifically, the CNN model in \cite{akhter2021abusive} had an approximate performance with most of the models in \cite{akhter2020automatic} on the Urdu dataset. However, its performance is much inferior to those in \cite{akhter2020automatic} on the Roman Urdu dataset, e.g., an $F_1$ score of 91.6\% by the CNN model in \cite{akhter2021abusive} against an $F_1$ score of 99.2\% by the LogitBoost model in \cite{akhter2020automatic}. The above performance inconsistencies of the state-of-the-art methods highlight the importance of our LLM-based approach as it demonstrates stable and consistent performance across different characteristics of data and languages. %which is critical in digital forensics and law enforcement.

%For the abusive datasets, the worst methods in [] [] for example, is xx xx. 

%quote F2 also in the paper.

%Our dataset is noiser and more challenging compared to Borj et al. \cite{borj2021detecting, rezaee2023detecting} because we did not eliminate non-English words that did not provide any special meaning or do not follow standard grammar, have many slang and emoticons.

%We consider our dataset is noisier and more challenging because did not remove non-English words or incorrect grammar expressions. 

%Talk about language, about imbalance, about the number of training data points. How it affects the performance or behaviors of training and test.

\section{Conclusion and Further Work}
\label{sec_con_fut}
The critical need for detecting online sexual predatory behaviours and abusive language cannot be overstated in today's digital age. As the internet becomes an integral part of our lives, it also provides a platform for various forms of misconduct and harm. Flagging and preventing online sexual predatory conversations and offensive texts are of paramount importance, not only to protect potential victims but also to contribute to the creation of a healthier online environment. In this paper, we have fine-tuned the pretrained Llama 2 7B-parameter LLM using the LoRA method for text classification with an interesting application in detecting online sexual predatory chat logs and abusive texts. Our proposed method is deemed a one-size-fits-all approach as it consistently delivered great results across various factors such as different languages, degrees of data imbalance, data sizes, and text lengths. We experimented our method on English, Roman Urdu and Urdu datasets and it demonstrated excellent performance in all tested languages. Even with a small number of non-English data samples used for fine-tuning, i.e., the Urdu dataset with only 1,736 training data points, the LLM-based approach outperformed most of the traditional text classification methods. It is worth noting that the Urdu language uses a completely different alphabet from English. The results of this study show a great potential of our approach, implying that it can work to detect online sexual predators and offensive texts not only in English but also in non-English languages.

As our method is automated and generic, it can be further applied to solving many other text classification problems in different fields such as \textit{cybersecurity} (e.g., phishing detection, malware detection based on textual characteristics), \textit{legal and compliance} (e.g., sorting legal documents, contracts, and agreements, or analyzing text data to ensure compliance with legal and regulatory requirements), \textit{social media and social listening} (e.g., fake news detection, topic trend analysis, or brand monitoring - tracking mentions and discussions about a brand or company on social media to gauge public perception), \textit{healthcare and medical research} (e.g., categorizing medical records or patient notes for research, billing, and patient care purposes), \textit{customer support} (e.g., analyzing customer feedback and reviews to identify common issues, concerns, and suggestions, or routing customer inquiries or support requests to the appropriate department or agent based on the content of the text), \textit{investment and finance} (e.g., sentiment analysis for trading, or assessing credit risk by analyzing textual data from loan applications and financial statements), and \textit{human resources} (e.g., analyzing employee feedback surveys and comments to understand job satisfaction and identify concerns, or resume screening - automatically categorizing job applicants' resumes based on their skills, qualifications, and experience). LLMs like Llama 2 models were trained on a huge text corpus, they certainly acquired significant world or general knowledge. Our fine-tuning approach implemented on top of these pretrained models will result in specialized knowledge for each of the aforementioned fields while preserving the general knowledge acquired previously during the pretraining process. It is therefore expected to demonstrate a great performance in solving various problems, like the results of sexual predatory chats and abusive language detection obtained in this study.

Another direction for future work would be to experiment our approach with bigger pretrained Llama 2 models such as 13B-, 34B- and 70B-parameter models. Unlike the 7B and 13B LLMs, the larger models, i.e., 34B and 70B, not only had more parameters, but also used the grouped-query attention method \cite{ainslie2023gqa} during the pretraining process to improve their inference scalability \cite{touvron2023llama}. These models are therefore worth being experimented with larger available fine-tuning datasets for tasks such as news categorization, social media monitoring, and research paper categorization. 

%educational technology
%Stable method when one model fits all kinds of data

\section*{Acknowledgment}
The authors would like to thank the Australian National Computational Infrastructure (NCI) for providing us with high-performance computing resources.

%Data size for non-english is small and the LLM was not pretrained using English, so this is encouraging results. 

%Pan12: to avoid subjective removal of conversations that have non-English words and incorrect grammars, we only perform these two preprocessing steps to ensure reproducibility of the results. With our method, we don’t need to manually remove non-English, grammar etc.

%Abusive texts can not be tolerate and can be criminal charges in some countries.
%Tables for data preparation
%Tables for results including all criteria printed, non-comparable will be  - - -

%Add keywords: cyber grooming, online chats, sexual predator, 

%Use two-sample testing to test training and test data to check same distribution or not? On embedding data

%Maybe scale to link with performance?
%Nonparametric testing, Permutation test
%Test imbalance of datasets as well and highlight the performance

%Blob dataset? %Non-translation invariant kernel

%Because of this limited feeding of pretrained Llama 2, we need more epochs of training on these abusive/toxic text data.

\bibliographystyle{IEEEtran}
\balance
\bibliography{BibTexFile}

\end{document}